\documentclass[runningheads]{llncs}
\usepackage{graphicx}
\usepackage{amsmath,amssymb} 
\usepackage{stmaryrd}
\usepackage{subfig}
\usepackage{url}
\usepackage{color}
\usepackage{multirow}
\usepackage{array}
\newcolumntype{P}[1]{>{\centering\arraybackslash}p{#1}}
\begin{document}
\pagestyle{headings}
\mainmatter
\def\ECCV18SubNumber{2717}  

\title{Zero-shot keyword spotting for visual speech recognition in-the-wild} 

\titlerunning{Zero-shot keyword spotting}
\authorrunning{T. Stafylakis \& G. Tzimiropoulos}

\author{Themos Stafylakis \and Georgios Tzimiropoulos}
\institute{Computer Vision Laboratory \\ University of Nottingham, U.K.
\email{\{themos.stafylakis,yorgos.tzimiropoulos\}@nottingham.ac.uk}
}
\maketitle

\begin{abstract}
Visual keyword spotting (KWS) is the problem of estimating whether a text query occurs in a given recording using only video information. This paper focuses on visual KWS for words unseen during training, a real-world, practical setting which so far has received no attention by the community. To this end, we devise an end-to-end architecture comprising (a) a state-of-the-art visual feature extractor based on spatiotemporal Residual Networks, (b) a grapheme-to-phoneme model based on sequence-to-sequence neural networks, and (c) a stack of recurrent neural networks which learn how to correlate visual features with the keyword representation. Different to prior works on KWS, which try to learn word representations merely from sequences of graphemes (i.e. letters), we propose the use of a grapheme-to-phoneme encoder-decoder model which learns how to map words to their pronunciation. We demonstrate that our system obtains very promising visual-only KWS results on the challenging LRS2 database, for keywords unseen during training. We also show that our system outperforms a baseline which addresses KWS via automatic speech recognition (ASR), while it drastically improves over other recently proposed ASR-free KWS methods.

\keywords{Visual keyword spotting \and visual speech recognition \and zero-shot learning}
\end{abstract}

\section{Introduction}

This paper addresses the problem of visual-only Automatic Speech Recognition (ASR) i.e. the problem of recognizing speech from video information only, in particular, from analyzing the spatiotemporal visual patterns induced by the mouth and lips movement. Visual ASR is a challenging research problem, with decent results being reported only recently thanks to the advent of Deep Learning and the collection of large and challenging datasets \cite{chung2017lipsent,chung2016lip,anina2015ouluvs2}. 

In particular, we focus on the problem of Keyword Spotting (KWS) i.e. the problem of finding occurrences of a text query among a set of recordings. In this work we consider only words, however the same architecture can be used for short phrases. Although the problem can be approached with standard ASR methods, recent works aim to address it with more direct and ``ASR-free'' methods \cite{audhkhasi2017end}. Moreover, such KWS approaches are in line with a recently emerged research direction in ASR (typically termed Acoustics-to-Word) where words are replacing phonemes, triphones or letters as basic recognition units \cite{audhkhasi2017,soltau2016neural}. 

\textbf{Motivation.} One of the main problems regarding the use of words as basic recognition units is the existence of Out-Of-Vocabulary (OOV) words, i.e. words for which the exact phonetic transcription is unknown, as well as words with very few or zero occurrences in the training set. This problem is far more exacerbated in the visual domain where collecting, annotating and distributing large datasets for fully supervised visual speech recognition is a very tedious process. To the best of our knowledge, this paper is the first attempt towards visual KWS under the zero-shot setting. 

\textbf{Relation to zero-shot learning.} Our approach shares certain similarities with zero-shot learning methods, e.g. for recognizing objects in images without training examples of the particular objects \cite{socher2013zero}. Different to \cite{socher2013zero}, where representations of the objects encode semantic relationships, we wish to learn word representations that encode merely their phonetic content. To this end, we propose to use a grapheme-to-phoneme (G2P) encoder-decoder model which learns how to map words (i.e. sequences of graphemes or simply letters) to their pronunciation (i.e. to sequences of phonemes)\footnote{For example, the phonetic transcription of the word ``finish''  is ``F IH1 N IH0 SH'', where the numerical values after the vowel ``IH'' indicate different levels of stretching.}. By training the G2P model using a training set of such pairs (i.e. words and their pronunciation), we obtain a fixed-length representation (embedding) for any word, including words not appearing in the phonetic dictionary or in the visual speech training set. 

The proposed system receives as input a video and a keyword and estimates whether the keyword is contained in the video. We use the LRS2 database to train a Recurrent Neural Network (Bidirectional Long Short-Term Memory, BiLSTM) that learns non-linear correlations between visual features and their corresponding keyword representation \cite{lrs2Website}. The backend of the network is modeling the probability that the video contains the keyword and provides an estimate of its position in the video sequence. The proposed system is trained end-to-end, without information about the keyword boundaries, and once trained it can spot any keyword, even those not included in the LRS2 training set.

In summary, our \textbf{contributions} are:
\begin{itemize}
\item
We are the first to study Query-by-Text visual KWS for words unseen during training.
\item
We devise an end-to-end architecture comprising (a) a state-of-the-art visual feature extractor based on spatiotemporal Residual Networks, (b) a G2P model based on sequence-to-sequence neural networks, and (c) a stack of recurrent neural networks that learn how to correlate visual features with the keyword representation.
\item
We demonstrate that our system obtains very promising visual-only KWS results on the challenging LRS2 database.
\end{itemize}

\section{Related Work}

\textbf{Visual ASR.} During the past few years, the interest in visual and audiovisual ASR has been revived. Research in the field is largely influenced by recent advances in audio-only ASR, as well as by the state-of-the-art in computer vision, mostly for extracting visual features. In \cite{assael2016lipnet}, CNN features are combined with Gated Recurrent Units (GRUs) in an end-to-end visual ASR architecture, capable of performing sentence-level visual ASR on a relatively easy dataset (GRID \cite{GRID}). Similarly to several recent end-to-end audio-based ASR approaches, CTC is deployed in order to circumvent the lack of temporal alignment between frames and annotation files \cite{graves2014towards,zweig2017advances}. In \cite{chung2017lipsent,chung2017profile}, the ``Listen, attend and spell'' (\cite{chan2016listen}) audio-only ASR architecture is adapted to the audiovisual domain, and tested on recently released in-the-wild audiovisual datasets. The architecture is an attentive encoder-decoder model with the decoder operating directly on letters (i.e. graphemes) rather than on phonemes or visemes (i.e. the visual analogues of phonemes \cite{bear2016decoding}). It deploys a VGG for extracting visual features and the audio and visual modalities are fused in the decoder. The model yields state-of-the-art results in audiovisual ASR. Other recent advances in visual and audiovisual ASR involve residual-LSTMs, adversarial domain-adaptation methods, use of self-attention layers (i.e. Transformer \cite{vaswani2017attention}), combinations of CTC and attention, gating neural networks, as well as novel fusion approaches \cite{Potamianos2017,petridis2018end,Wand2017,afouras2018deep,xu2018lcanet,sterpu2018can,tao2018gating,mroueh2015deep}.

\textbf{Words as recognition units.} The general tendency in deep learning towards end-to-end architectures, together with the challenge of simplifying the fairly complex traditional ASR paradigm, has resulted into a new research direction of using words directly as recognition units. In \cite{bengio2014word}, an acoustic deep architecture is introduced, which models words by projecting them onto a continuous embedding space. In this embedding space, words that sound alike are nearby in the Euclidean sense, differentiating it from others word embedding spaces where distances correspond to syntactic and semantic relations \cite{pennington2014glove,mikolov2013distributed}. In \cite{audhkhasi2017,soltau2016neural}, two CTC-based ASR architectures are introduced, where CTC maps directly acoustics features to words. The experiments show that CTC word models can outperform state-of-the-art baselines that make use of context-dependent triphones as recognition units, phonetic dictionaries and language models. 

In the problem of audio-based KWS, end-to-end word-based approaches have also emerged. In \cite{palaz2016jointly}, the authors introduce a KWS system based on sequence training, composed of a CNN for acoustic modeling and an aggregation stage, which aggregates the frame-level scores into a sequence-level score for words. However, the system is limited to words seen during training, since it merely associates each word with a label (i.e. one-hot vector) without considering them as sequences of characters. Other recent works aim to spot specific keywords used to activate voice assistant systems \cite{sun2017compressed,sun2015model,chen2014small}. The application of BiLSTMs on KWS was first proposed in \cite{fernandez2007application}. The architecture is capable of spotting at least a limited set of keywords, having a softmax output layer with as many output units as keywords, and a CTC loss for training. More recently, the authors in \cite{audhkhasi2017end} propose an audio-only KWS system capable of generalizing to unseen words, using a CNN/RNN to autoencode sequences of graphemes (corresponding to words or short phrases) into fixed-length representation vectors. The extracted representations, together with audio-feature representations extracted with an acoustic autoencoder are passed to a feed-forward neural network which is trained to predict whether the keyword occurs in the utterance or not. Although this audio-only approach shares certain conceptual similarities with ours, the implementations are different in several ways. Our approach deploys a Grapheme-to-Phoneme model to learn keyword representations, it does not make use of autoencoders for extracting representations of visual sequences, and more importantly it learns how to correlate visual information with keywords from low-level visual features rather than from video-level representations. 

The authors in \cite{jha2018word} recently proposed a visual KWS approach using words as recognition units. They deploy the ResNet feature extractor with us (proposed by our team in \cite{Stafy2017,stafylakis2017deep} and trained on LRW \cite{chung2016lip}) and they demonstrate the capacity of their network in spotting occurrences of the $N_w = 500$ words in LRW \cite{lrwWebsite}. The bottleneck of their method is the word representation (each word corresponds to a label, without considering words as sequences of graphemes). Such an unstructured word representation may perform well on closed-set word identification/detection tasks, but prevents the method from generalizing to words unseen during training. 

\textbf{Zero-shot learning.} Analogies can be drawn between KWS with unseen words and zero-shot learning for detecting new classes, such as objects or animals. KWS with unseen words is essentially a zero-shot learning problem, where attributes (letters) are shared between classes (words) so that the knowledge learned from seen classes is transfered to unseen ones \cite{xian2017zero}. Moreover, similarly to a typical zero-shot learning training set-up where bounding boxes of the objects of interest are not given, a KWS training algorithm knows only whether or not a particular word is uttered in a given training video, without having information about the exact time interval. For these reasons, zero-shot learning methods which e.g. learn mappings from an image feature space to a semantic space (\cite{frome2013devise,akata2016label}) are pertinent to our method. Finally, recent methods in action recognition using a representation vector to encode e.g. 3D human-skeleton sequences also exhibit certain similarities with our method \cite{Mahasseni_2016_CVPR}.

\section{Proposed Method}
\subsection{System overview} 

Our system is composed of four different modules. The first module is a visual feature extractor, which receives as input the image frame sequence (assuming a face detector has already been applied, as in LRS2) and outputs features. A spatiotemporal Residual Network is used for this purpose, which has shown remarkable performance in word-level visual ASR \cite{Stafy2017,stafylakis2017deep}.

The second module of the architecture receives as input a user-defined keyword (or more generally a text query) and outputs a fixed-length representation of the keyword in $\mathbb{R}^{d_e}$. This mapping is learned by a grapheme-to-phoneme (G2P \cite{yao2015sequence}) model, which is a sequence-to-sequence neural network with two RNNs playing the roles of encoder and decoder (similarly to \cite{sutskever2014sequence}). The two RNNs interact with each other via the last hidden state of the encoder, which is used by the decoder in order to initialize its own hidden state. We claim that this representation is a good choice for extracting word representations, since (a) it contains information about its pronunciation without requiring the phonetic transcription during evaluation, and (b) it generalizes to words unseen during training, provided that the G2P is trained with a sufficiently large vocabulary. 

The third module is where the visual features with the keyword representation are combined and non-linear correlations between them are learned. It is implemented by a stack of bidirectional LSTMs, which receives as input the sequence of feature vectors and concatenates each such vector with the word representation vector. 

Finally, the forth module is the backend classifier and localizer, whose aims are (a) to estimate whether or not the query occurs in the video, and (b) to provide us with an estimate of its position in the video. Note that we do not train the network with information about the time intervals keywords occur. The only supervision used during training is a binary label indicating whether or not the keyword occurs in the video, together with the grapheme and phoneme sequences of the keyword.

\begin{figure}[!htbp]
\centering
\includegraphics[width=4.20in]{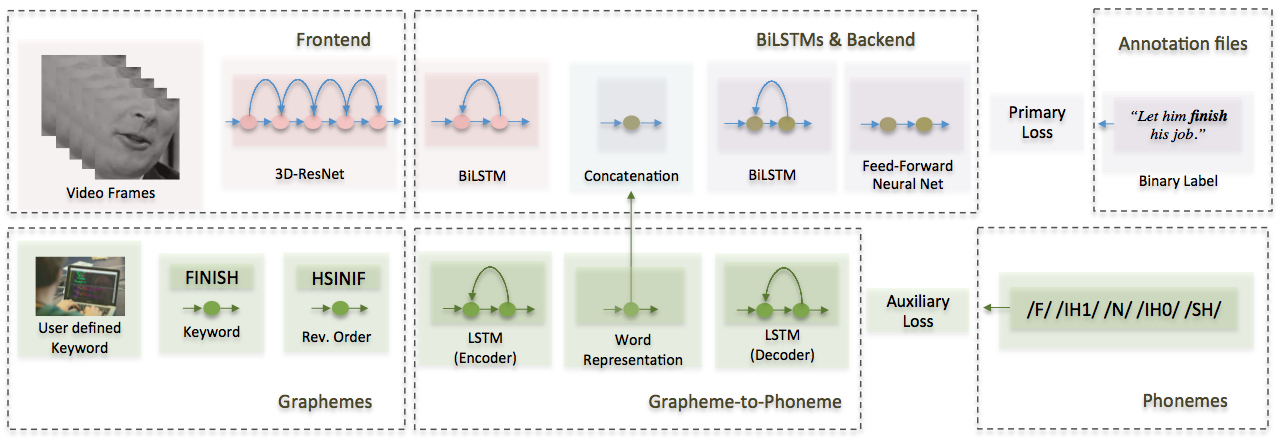}
\caption{The block-diagram of the proposed KWS system.}
\label{fig:model}
\end{figure}

The basic building blocks of the model are depicted in Fig. \ref{fig:model}.

\subsection{Modeling visual patterns using spatiotemporal ResNet}
\label{ss:pretrain}
The front-end of the network is an 18-layer Residual Network (ResNet), which has shown very good performance on LRW \cite{Stafy2017} \cite{he2016identity} as well as on LRS2 \cite{afouras2018deep}. It has been verified that CNN features encoding spatiotemporal information in their first layers yield much better performance in lipreading, even when combined with deep LSTMs or GRUs in the backend \cite{Stafy2017,assael2016lipnet,chung2017profile}. For this reason, we replace the first 2D convolutional, batch-normalization and max-pooling layers of the ResNet with their 3D counterparts. The temporal size of the kernel is set equal to $T_{r} = 5$, and therefore each ResNet feature is extracted over a window of 0.2s (assuming 25fps). The temporal stride is equal to 1, since any reduction of time resolution is undesired at this stage. Finally, the average pooling layer of the ResNet output (found e.g. in ImageNet versions of ResNet \cite{he2016identity}) is replaced with a fully connected layer. Overall, the spatiotemporal ResNet implements a function $\mathbf{x}_t = f_{r}\left(\left[ \mathbf{I}_{t-2},\mathbf{I}_{t-1},\mathbf{I}_{t},\mathbf{I}_{t+1},\mathbf{I}_{t+2}   \right], \mathbf{W}_r \right)$, where $\mathbf{W}_r$ denotes the parameters of the ResNet and $\mathbf{I}_{t}$ the (grayscale and cropped) frames at time $t$. 

We use a pretrained model on LRW which we fine-tune on the pretrain set of LRS2 using closed-set word identification. The pretrain set of LRS2 is useful for this purpose, not merely due to the large number of utterances it contains, but also sue to its more detailed annotation files, which contain information about the (estimated) time each word begins and ends. Word boundaries permits us to excerpt fixed-duration video segments containing specific words and essentially mimic the LRW set-up. To this end, we select the 2000 most frequently appearing words containing at least 4 phonemes and we extract frame sequences of 1.5sec duration, having the target word in the center. The backend is a 2-layer LSTM (jointly pretrained on LRW) which we remove once the training is completed.

\textbf{Preprocessing.} The frames in LRS2 are already cropped according the bounding box extracted by face detector and tracker \cite{chung2017lipsent,chung2016lip}. We crop the frames further with a fixed set of coefficients $C_{crop}= \left[15,46,145,125 \right]$, we resize them to $122\times 122$, and we finally feed the ResNet with frames of size $112 \times 112$, after applying random cropping in training (for data augmentation), and fixed central cropping in testing, as in \cite{Stafy2017}.

\subsection{Grapheme-to-phoneme models for encoding keywords}
Grapheme-to-phoneme (G2P) models are extensively used in speech technologies in order to learn a mapping $\mathbf{G} \mapsto \mathbf{P}$ from sequences of graphemes $\mathbf{G} \in \mathbb{G}$ to sequences of phonemes $\mathbf{P} \in \mathbb{P}$. Such models are typically trained in a supervised fashion, using a phonetic dictionary, such as the CMU dictionary (for English). The number of different phonemes in the CMU dictionary is equal to $N_{phn} = 69$, with each vowel contributing more than one phoneme, due to the variable level of stretching. The effectiveness of a G2P model is measured by its generalizability, i.e. by its capacity in estimating the correct pronunciation(s) of words unseen during training. 

Sequence-to-sequence neural networks have recently shown their strength in addressing this problem \cite{yao2015sequence}. In a sequence-to-sequence G2P model, both sequences are typically modeled by an RNN, such as an LSTM or a GRU. The first RNN is a function $\mathbf{r} = f_{e}(\mathbf{G},\mathbf{W}_{e})$ parametrized by $\mathbf{W}_{e}$, which encodes the grapheme sequence in a fixed-size representation $\mathbf{r}|\mathbf{G}$, where $\mathbf{r} \in \mathbb{R}^{d_r}$, while the second RNN estimates the phoneme sequence $\mathbf{\hat{P}} = f_{d}(\mathbf{r},\mathbf{W}_{d})$. The representation vector is typically defined as the output of the last step, i.e. once the RNN has seen the whole grapheme sequence. 

Our implementation of G2P involves two unidirectional LSTMs with hidden size equal to $d_{l} = 64$. Similarly to sequence-to-sequence models for machine translation (e.g. \cite{sutskever2014sequence}), the encoder receives as input the (reversed) sequence of graphemes and the decoder received $\mathbf{c}_{e,T}$ and the output $\mathbf{h}_{e,T}$ from the encoder (corresponding to the last time step $t=T$) to initialize its own state, denoted by $\mathbf{c}_{d,0}$ and $\mathbf{h}_{d,0}$. To extract the word representation $\mathbf{r}$, we first concatenate the two vectors, we then project them to $\mathbb{R}^{d_{r}}$ to obtain $\mathbf{r}$ and finally we re-project them back to $\mathbb{R}^{2d_{l}}$, i.e. $\left[ \mathbf{c}_{e,T}^{t},\mathbf{h}_{e,T}^{t}\right]^{t} \mapsto \mathbf{r} \mapsto \left[ \mathbf{c}_{d,0}^{t},\mathbf{h}_{d,0}^{t}\right]^{t}$, where $\mathbf{x}^t$ denotes the transpose of $\mathbf{x}$. For the projections we use two linear layers with square matrices (since $d_r = 2d_l$), while biases are omitted for having a more compact notation.

The G2P model is trained by minimizing the cross-entropy (CE) between the true $\mathbf{P^{*}}$ and posterior probability over sequences $P(\mathbf{P}_t|\mathbf{G})$, averaged across time steps, i.e.
\begin{equation}
L_{w}\left(\mathbf{P}^{*},\mathbf{G}\right) = \frac{1}{T}\sum_{t=1}^{T} \textrm{CE}\left(\mathbf{P}^{*}_t, P(\mathbf{P}_t|\mathbf{G})\right). 
\end{equation}
Since the G2P model is trained with back-propagation, its loss function can be added as auxiliary loss to the primary KWS loss function and the overall architecture can be trained jointly. Joint training is highly desired, as it enforces the encoder to learn representations that are optimal not merely for decoding, but for our primary task, too. 

During evaluation, the mapping $\mathbf{G} \mapsto \mathbf{z}$  learned by the encoder is all that is required, and therefore the decoder $f_{dec}(\cdot,\mathbf{W}_{dec})$ and the true pronunciation $\mathbf{P}^{*}$ are not required for KWS. 

\subsection{Stack of BiLSTMs, binary classifier and loss function}

The backend of the model receives the sequence of visual features $\mathbf{X} = \{\mathbf{x}_{t} \}_{t=1}^{T}$ of a video and the word representation vector $\mathbf{r}$ and estimates whether the keyword is uttered by the speaker. 

\textbf{Capturing correlations with BiLSTMs.} LSTMs have exceptional capacity in modeling long-term correlations between input vectors, as well as correlations between different entries of the input vectors, due to the expressive power of their gating mechanism which controls the memory cell and the output \cite{hochreiter1997long}. We use two bidirectional LSTM (BiLSTM), with the first BiLSTM merely applying a transformation of the feature sequence $\mathbf{X} \mapsto \mathbf{Y}$, i.e. 
\begin{equation}
\left[\left(\overset{\shortrightarrow}{\mathbf{h}}_{t}, \overset{\shortrightarrow}{\mathbf{c}}_{t} \right), \left(\overset{\shortleftarrow}{\mathbf{h}}_{t}, \overset{\shortleftarrow}{\mathbf{c}}_{t} \right)\right] = \left[\overset{\shortrightarrow}{f}_{l_0} \left( \mathbf{x} _{t}, \overset{\shortrightarrow}{\mathbf{h}}_{t-1} , \overset{\shortrightarrow}{\mathbf{c}}_{t-1} \right), \overset{\shortleftarrow}{f}_{l_0} \left( \mathbf{x} _{t}, \overset{\shortleftarrow}{\mathbf{h}}_{t+1} , \overset{\shortleftarrow}{\mathbf{c}}_{t+1} \right)\right]
\end{equation} 

and
\begin{equation}
\mathbf{y}_t = \mathbf{W}^t_{l_0} \left[\overset{\shortrightarrow}{\mathbf{h}}^{t}_{t},\overset{\shortleftarrow}{\mathbf{h}}^{t}_{t} \right]^{t}
\end{equation}
where $\mathbf{W}_{l_0}$ is a linear layer of size $(2d_{v}, d_{v})$, $\overset{\shortrightarrow}{f}_{l_0}$ and $\overset{\shortleftarrow}{f}_{l_0}$ are functions corresponding to the forward and backward LSTM models (the dependence on their parameters is kept implicit), while $d_v = 256$.
The input vectors $\mathbf{X}$ are batch-normalized, and dropouts with $p=0.2$ are applied by repeating the same dropout mask for all feature vectors of the same sequence \cite{DropoutsRNN}. The outputs vectors of the first BiLSTM are concatenated with the word representation vector to obtain $\mathbf{y}^{+}_{t} = \left[ \mathbf{y}_{t}^{t}, \mathbf{r}^{t} \right]^t$. After applying batch-normalization to $\mathbf{y}^{+}_{t}$, we pass them as input to the second BiLSTM, with equations defined as above, resulting in a sequence of output vectors denoted by $\mathbf{Z} = \{\mathbf{z}_t \}_{t=1}^{T}$, where $\mathbf{z}_t \in \mathbb{R}^{d_v}$. 

Note the equivalence between the proposed frame-level concatenation and keyword-based model adaptation. We may consider $\mathbf{r}$ as a means to adapt the biases of the linear layers in the three gates and the input to the cell, in such a way so that the activations of its neurons fire only over subsequences in $\mathbf{Z}$ that correspond to the keyword encoded in $\mathbf{r}$. 

\textbf{Feed-Forward Classifier for network initialization.} For the first few epochs, we use a simple feed-forward classifier, which we subsequently replace with a BiLSTM backend discussed below.  
The outputs of the BiLSTM stack are projected to a linear layer $(d_v,d_v /2)$ and are passed to a non-linearity (Leaky Rectified Linear Units, denoted by LReLU) to filter-out those entries with negative values, followed by a summation operator to aggregate over the temporal dimension, i.e. $\mathbf{v} = \sum_{t=1}^T \textrm{LReLU}(\mathbf{W}^t\mathbf{z}_t)$. After applying dropouts to $\mathbf{v}$ we project them to a linear layer $(d_v/2, d_v/4)$ and we apply again a LReLU layer. Finally, we apply a linear layer to drop the size from $d_v/4$ to 1 and a Sigmoid layer, with which we model the posterior probabilities that the video contains the keyword or not, i.e. $P\left(l|\{\mathbf{I}\}_{t=1}^{T}, \mathbf{G} \right)$, where $l \in \{0,1\}$ the binary indicator variable and $l^{*}$ its true value. 

\begin{figure}[!htbp]
\centering
\includegraphics[width=4.00in]{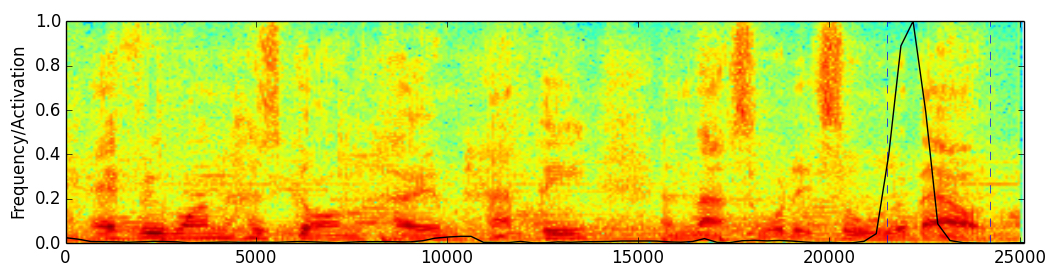}
\caption{Localization of the keyword {\bf about} in the phrase ``Everyone has gone home happy and that's what it's all {\bf about}''. The keyword boundaries are depicted with two vertical lines over the log-spectrogram.}
\label{fig:local}
\end{figure}

\textbf{BiLSTM Classifier and keyword localization.} Once the network with the Feed-Forward Classifier is trained, we replace it with an BiLSTM classifier. The latter does not aggregate over the temporal dimension, because it aims to jointly (a) estimate the posterior probability that the video contains the keyword, and (b) locate the time step that the keyword occurs. Recall that the network is trained without information about the actual time intervals the keyword occurs. Nevertheless, an approximate position of the keyword can still be estimated, even from the output of the BiLSTM stack. As Fig. \ref{fig:local} shows, the average activation of the input of the BiLSTM classifier (after applying the linear layer and ReLU) exhibits a peak, typically within the keyword boundaries. The BiLSTM Classifier aims to model this property, by applying $\textrm{max}(\cdot)$ and $\textrm{argmax}(\cdot)$ in order to estimate the posterior that the keyword occurs and localize the keyword, respectively. 
More analytically, the BiLSTM Classifier receives the output features of the BiLSTM stack and it passes them to a linear layer $\mathbf{W}$ of size $(d_v,d_s)$ where $d_l=16$, and to a LReLU, i.e. $\mathbf{s}_t = \textrm{LReLU}(\mathbf{W}^t\mathbf{z}_t)$. The BiLSTM is then applied on the sequence, followed by a linear layer (which drops the dimension from $2d_s$ to 1, i.e. a vector $\mathbf{w}$ and a bias $b$), the $\textrm{max}(\cdot)$ and finally the sigmoid $\sigma(\cdot)$ from which we estimate the posterior. More formally, $\mathbf{H} = \textrm{BiLSTM}(\mathbf{S})$, $y_t = \mathbf{w}^t\mathbf{h}_t + b$ and $p = \sigma(\max(\mathbf{y}))$, $\hat{t}=\textrm{argmax}(\mathbf{y})$ where $\mathbf{S} = \{\mathbf{s}_t\}_{t=1}^T$, $\mathbf{H} = \{\mathbf{h}_t\}_{t=1}^T$, $\mathbf{y} = \{y_t\}_{t=1}^T$, $p=P\left(l=1|\{\mathbf{I}_t\}_{t=1}^{T}, \mathbf{G} \right)$ (i.e. the posterior that the keyword defined by $\mathbf{G}$ occurs in the frame sequence $\{\mathbf{I}_t\}_{t=1}^{T}$), and $\hat{t}$ is the time step where the maximum occurs, and should be somewhere within the actual keyword boundaries. Note that we did not succeed in training the network with the BiLSTM Classifier from scratch, probably due to the $\textrm{max}(\cdot)$ operator.  
 
\textbf{Loss for joint training.} The primary loss is defined as:
\begin{equation}
L_{v}\left(l^{*},\left[\{\mathbf{I}_t\}_{t=1}^{T}, \mathbf{G} \right]\right) = \textrm{CE}\left(l^{*},P\left(l|\{\mathbf{I}_t\}_{t=1}^{T},  \mathbf{G} \right)\right),
\end{equation}
while the whole model is trained jointly by minimizing a weighted summation of the primary and auxiliary losses, i.e.
\begin{equation}
\label{eq:loss}
L\left(\left[ l^{*},\mathbf{P^{*}} \right],\left[\{\mathbf{I}_t\}_{t=1}^{T}, \mathbf{G} \right]\right) = L_{v}\left(l^{*},\left[\{\mathbf{I}_t\}_{t=1}^{T}, \mathbf{G} \right]\right) + \alpha_{w} L_{w}\left(\mathbf{P}^{*},\mathbf{G}\right), 
\end{equation}
where $\alpha_{w}$ a scalar for balancing the two losses. It is worth noting that the representation vectors $\mathbf{r}$ and the encoder's parameters receive gradients from both loss functions, via the decoder of the G2P model and the LSTM backend. Contrarily, the decoder and the binary classifier receive gradients only from $L_{w}(\cdot,\cdot)$ and $L_{v}(\cdot,\cdot)$, respectively.  

\section{Training the model}

In this section we describe our recipe for training the model. We explain how we partition the data, how we create minibatches, and we give details about the optimization parameters. 

\subsection{LRS2 and CMU Dictionary partitions}

We use the official partition of the LRS2 into pretrain, train, validation and test set. The KWS network is trained on pretrain and train sets. The pretrain set is also used to fine-tune the ResNet, as we discuss in Section \ref{ss:pretrain}. The G2P model is trained from scratch and jointly with the whole KWS network. LRS2 contains about 145K videos of spoken sentences from BBC TV (96K in pretrain, 46K in train, 1082 in validation, and 1243 in test set). The number of frames per video in the test set varies between 15 and 145.

In terms of keywords, we randomly partition the CMU phonetic dictionary into train, validation and test words (corresponding to 0.75, 0.05 and 0.20, respectively), while words with less that $n_p=4$ phonemes are removed. Finally, we add to the test set of the dictionary those words we initially assigned to the training and validation sets that do not occur in the LRS2 pretrain or train sets, since they are not used in any way during training. 

\subsection{Minibatches, training sets and backend}
\label{ss:minibatches}
Minibatches for training the KWS model should contain both positive and negative examples, i.e. pairs of videos and keywords where each pair is considered as positive when the video contains the corresponding keyword and negative otherwise. Epochs and minibatches are defined based on the videos, i.e. each epoch contains all videos of the train and pretrain set of LRS2, partitioned into minibatches. The list of keywords in each minibatch is created by all words occurring in the minibatch belonging to the training set of CMU dictionary and having at least $n_p$ number of phonemes. At each minibatch, each video is paired with (a) all its keywords (positive pairs) and (b) an equal number of other randomly chosen keywords from the list (negative pairs). This way we ensure that each video has equal number of positive and negative examples. At each epoch we shuffle the videos in order to create new negative pairs. By feeding the algorithm with the same set of videos and keywords under different binary labels in each minibatch, we enforce it to capture the correlations between videos and words, instead of attempting to correlate the binary label with certain keywords or with irrelevant aspects of specific videos. 

For the first 20 epochs we use (a) only the train set of LRS2 (because it contains shorter utterances and much fewer labeling errors compared to the pretrain), (b) $n_p=4$ and $\alpha_w = 1.0$ (i.e. minimum number of phonemes and weight of auxiliary loss, respectively), and (c) the simple feed-forward backend. After the 20th epoch (a) we add the pretrain set, (b) we set $n_p=6$ and $\alpha_w = 0.1$, and (c) we replace the backend with the BiLSTM-based (all network parameters but those of the backend are kept frozen during the 21st epoch). 
\subsection{Optimization}
The loss function in eq. (\ref{eq:loss}) is optimized with backpropagation using the Adam optimizer \cite{Adam}. The number of epochs is 100, the initial learning rate is $2\times 10^{-3}$ and we drop it by a factor of $2$ every 20 epochs. The best model is chosen based on the performance on the validation set. The implementation is based on PyTorch and the code together with pretrained models and ResNet features will be released soon. The number of videos in each minibatch is 40, however, as explained in Section \ref{ss:minibatches}, we create multiple training examples per video (equal to twice the number of training keywords it contains). Finally, the ResNet is optimized with the configuration suggested in \cite{Stafy2017}.

\section{Experiments}
\label{sec:Experiments}
We present here the experimental set-up, the metrics we use and the results we obtain using the proposed KWS model. Moreover, we report baseline results using (a) a visual ASR model with a hybrid CTC/attention architecture, and (b) an implementation of the ASR-free KWS method recently proposed in \cite{audhkhasi2017end}.  
\subsection{Evaluation metrics and keyword selection}

KWS is essentially a detection problem, and in such problems the optimal threshold is application-dependent, typically determined by the desired balance between the false alarm rate (FAR) and the missed detection rate (MDR). Our primary error metric is the Equal Error Rate (EER), defined as the FAR (or MDR) when the threshold is set so that the two rates are equal. We also report MDR for certain low values of FAR (and vice versa) as well as FAR vs. MDR curves. 

Apart from EER, FAR and MDR we evaluate the performance based on ranking measures. More specifically, for each text query (i.e. keyword) we report the percentage of times the score of a video containing the query is within the Top-$N$ scores, where $N \in \{1,2,4,8\}$. Since a query $q$ may occur in more than one videos, a positive pair with score $s_{q,v'}$ is considered as Top-$N$ if the number of negative pairs associated with the given query $q$ with score higher than $s_{q,v'}$ is less than $N$, i.e. if $|\{q,v| l_{q,v} = 0, s_{q,v}> s_{q,v'}\}| < N$. 

The evaluation is performed by creating a list of single-word queries, containing all words appearing in the test utterances and having at least 6 phonemes. Keywords appearing in the training and development sets are removed from the list. The final number of queries in the list is $N_q=635$. Each query is scored against all $N_{test}= 1243$ test videos, so the number of all pairs is $N_{q} N_{test}= 789305$. The number of positive pairs is $N_p = |\{q,v| l_{q,v} = 1\}|=873$, and $N_p>N_q$ because some keywords appear in more than one videos.   
\subsection{Baseline and proposed networks}
\textbf{CTC/Attention Hybrid ASR model.} We present here our baseline obtained with a ASR-based model. We use the same ResNet features but a deeper (4-layer) and wider (320-unit) BiLSTM. The implementation is based on the open-source ESPnet Python toolkit presented in \cite{watanabe2018espnet} using the hybrid CTC/attention character-level network introduced in \cite{watanabe2017hybrid}. The system is trained on the pretrain and train sets of LRS2, while for training the language model we also use the Librispeech corpus \cite{panayotov2015librispeech}. The network attains $\textrm{WER} = 71.4\%$ on the LRS2 test set. In decoding, we use the single step decoder beam search (proposed in \cite{watanabe2017hybrid}) with $|H|=40$ number of decoding hypotheses $h \in H$. Similarly to \cite{miller2007rapid}, instead of searching for the keyword only on the best decoding hypothesis we approximate the posterior probability that a keyword $q$ occurs in the video $v$ with feature sequence $\mathbf{X}$ as follows:
\begin{equation}
\label{eq:ASRposterior}
P(l=1|\mathbf{q,X}) = \sum_{h \in H}{\mathbf{1}_{[q \in h ]} P(h|\mathbf{X})},
\end{equation}
\begin{equation}
\label{eq:ASRposterior2}
P(h|\mathbf{X}) \approx \frac{\exp({s_h/c})}{\sum_{h' \in H}{\exp({s_{h'}/c})}},
\end{equation}
where $\mathbf{1}_{[q \in h ]}$ is the indicator function that the decoding hypothesis $h$ contains $q$, $s_h$ is the score (log-likelihood) of hypothesis $h$ (combining CTC and attention \cite{watanabe2017hybrid}) and $c=5.0$ is a fudge factor optimized in the validation set.

\textbf{Baseline with video embeddings.} We implement an ASR-free method that is very close to \cite{audhkhasi2017end} proposed for audio-based KWS. Different to \cite{audhkhasi2017end} we use our LSTM-based encoder-decoder instead of the proposed CNN-based. A video embedding is extracted from the whole utterance, is concatenated with the word representation and fed to a feed-forward binary classifier as in \cite{audhkhasi2017end}. This network is useful in order to emphasize the effectiveness of our frame-level concatenation. 

\textbf{Proposed Network and alternative Encoder-Decoder losses.} To assess the effectiveness of the proposed G2P training method, we examine 3 alternative strategies: (a) The encoder receives gradients merely from the decoder, which is equivalent to training a G2P network separately, using only the words appearing in the training set. (b) The network has no decoder, auxiliary loss or phoneme-based supervision, i.e. the encoder is trained by minimizing the primary loss only. (c) A Grapheme-to-Grapheme (G2G) network is used instead of a G2P. The advantage of this approach over G2P is that it does not require a pronunciation dictionary, i.e. it requires less supervision. The advantage over the second approach is the use of the auxiliary loss (over graphemes instead of phonemes), which acts as a regularizer.

\subsection{Experimental Results on LRS2.} 
\begin{figure}[t]
    \centering
        \subfloat[]{\includegraphics[width=0.49\columnwidth]{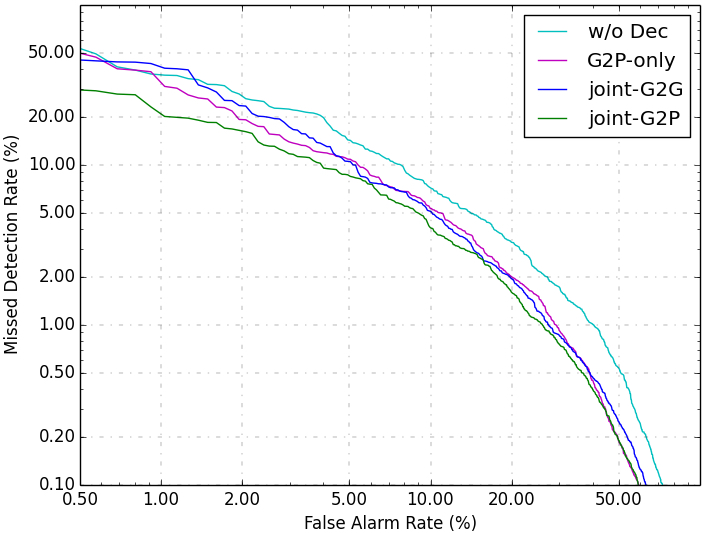}}
        \hfill
        \subfloat[]{\includegraphics[width=0.49\columnwidth]{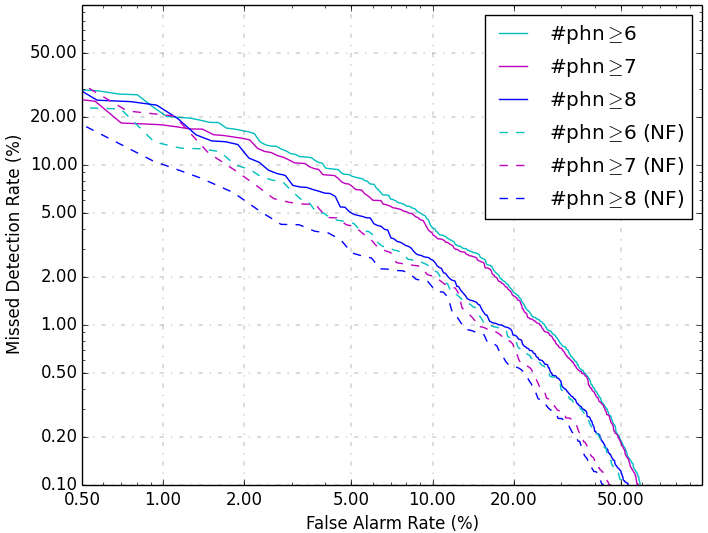}}        	
            \caption{FAR-MDR tradeoff. (a) Comparison between configurations of the proposed network. (b) The effect of the minimum number of phonemes per keyword and the camera view in the performance attained by Joint-G2P. As expected, longer keywords and near frontal (NF) view yield better results.}
    \label{fig:DET}
\end{figure}

Our first set of results based on the detection metrics are given in Table \ref{tbl.ResCTCLRS}. We observe that all variants of the proposed network attain much better performance compared to video embeddings. Clearly, video-level representations cannot retain the fine-grained information required to spot individual words. Our best network is the proposed Joint-G2P network (i.e. KWS network jointly trained with G2P), while the degradation of the network when graphemes are used as targets in the auxiliary loss (Joint-G2G) underlines the benefits from using phonetic supervision. Nevertheless, the degradation is relatively small, showing that the proposed architecture is capable of learning basic pronunciation rules even without phonetic supervision. Finally, the variant without a decoder during training is inferior to all other variants (including Joint-G2G), showing the regularization capacity of the decoder. The FAR-MDR tradeoff curves are depicted in Fig. \ref{fig:DET}(a), obtained by shifting the decision threshold which we apply to the output of the network. The curves show that the proposed architecture with G2P and joint training is superior to all others examined and in all operating points. Finally, we omit results obtained with the ASR-based model as the scoring rule described in eq. (\ref{eq:ASRposterior})-(\ref{eq:ASRposterior2}) is inadequate for measuring EER. The model yields very low FAR ($\approx 0.2\%$) at the cost of very high MDR ($\approx 63\%$) of all reasonable operating points.

\begin{table}[!htbp]
\caption{Equal-Error, False Alarm and Missed Detection Rates}
\centering
\begin{tabular}{P{2.5cm} P{1.8cm} P{1.8cm} P{1.8cm} P{1.8cm} P{1.8cm} P{1.8cm}} 
\hline
Network & EER & MDR$_{\textrm{FAR}=5\%}$ & MDR$_{\textrm{FAR}=1\%}$ &  FAR$_{\textrm{MDR}=5\%}$ & FAR$_{\textrm{MDR}=1\%}$ \\ [0.5ex] 
\hline
Video Embed. & 32.09\% & 77.32\% & 92.67\% & 66.76\% & 83.57\% \\ 
\hline
Prop. w/o Dec.   & 8.46\% & 14.09\% & 40.32\% & 14.25\% & 36.43\% \\ 
\hline
Prop. G2P-only & 7.22\% & 10.88\% & 29.21\% & 10.85\% & 30.99\%  \\ 
\hline
Prop. Joint-G2G & 7.26\% & 10.08\% & 27.38\% & 10.51\% & 40.26\%  \\ 
\hline
Prop. Joint-G2P & \bf{6.46\%} & \bf{8.93\%}  & \bf{26.00\%} & \bf{8.48\%} & \bf{20.11\%} \\ 
\hline
\end{tabular}
\label{tbl.ResCTCLRS}
\end{table}

\textbf{Length of keywords and camera view.} We are also interested in examining the extent to which the length of the keyword affects the performance. To this end, we increase the minimum number of phonemes from $n_p = 6$ to 7 and 8. Moreover, we evaluate the network only on those videos labeled as Near-Frontal (NF) view, by removing those labeled as Multi-View (the labeling is given in the annotation files of LRS2). The results are plotted in Fig. \ref{fig:DET}(b). As expected, the longer the keywords, the lower the error rates. Moreover, the performance is better when only NF views are considered.    

\textbf{Ranking measures and localization accuracy.} We measure here the percentage of times videos containing the query are in the top-$N$ scores. The results are given in Table \ref{tbl.Rank}. As we observe, our best system scores Top-1 equal to 34.14\% meaning that in about 1 out of 3 queries, the video containing the query is ranked first amongst the $N_{test} = 1243$ videos. Moreover, in 2 out of 3 queries the video containing the query is amongst the Top-8. The other training strategies perform well, too, especially the one where the encoder is trained merely with the auxiliary loss (G2P-only). The ranking measures attained by the Video-Embedding method are very bad so we omit them. The ASR-based system attains relatively high Top-1 score, however the rest of the scores are rather poor. We should emphasize though that other ASR-based KWS methods exist for approximating the posterior of a keyword occurrence, e.g. using explicit keyword lattices \cite{zhuang2016unrestricted}, instead of using the set of decoding hypotheses $H$ created by the beam search in eq. (\ref{eq:ASRposterior})-(\ref{eq:ASRposterior2}).

Finally, we report the localization accuracy for all versions of the proposed network, defined as the percentage of times the estimated location $\hat{t}$ lies within the keyword boundaries ($\pm 2$ frames). The reference word boundaries are estimated by applying forced alignment between the audio and the actual text. We observe that although the algorithm is trained without any information about the location of the keywords, it can still provide a very precise estimate of the location of the keyword in the vast majority of cases. 

\begin{table}[!htbp]
\caption{Ranking results showing the rate by which the video sequence containing the keyword is amongst the top-$N$ scores. Localization accuracy is also provided}
\centering
\begin{tabular}{P{2.5cm} P{1.8cm} P{1.8cm} P{1.8cm} P{1.8cm} P{1.8cm} P{1.8cm}} 
\hline
Network & Top-1 & Top-2 & Top-4 & Top-8 &  Local. Acc.\\ [0.5ex] 
\hline
ASR-based & 24.51\% & 31.39\% & 33.51\% & 37.57\% & - \\
\hline
Prop. w/o Dec   & 23.71\% & 33.68\% & 43.99\% & 55.90\% & 96.20\%\\ 
\hline
Prop. G2P-only & \bf{34.14\%} & 46.28\% & \bf{57.16\%} & 65.75\%  & 97.39\% \\ 
\hline
Prop. Joint-G2G & 31.16\% & 43.07\% & 54.98\% & 65.75\% & \bf{97.86}\% \\ 
\hline
Prop. Joint-G2P & \bf{34.14\%} & \bf{46.96\%} & 57.04\% & \bf{67.70\%} & 96.67\%  \\ 
\hline
\end{tabular}
\label{tbl.Rank}
\end{table}


\section{Conclusions}
We proposed an architecture for visual-only KWS with text queries. Rather than using subword units (e.g. phonemes, visemes) as main recognition units, we followed the direction of modeling words directly. Contrary to other word-based approaches, which treat words merely as classes defined by a label (e.g. \cite{stafylakis2017deep}), we inject into the model a word representation extracted by a grapheme-to-phoneme model. This zero-shot learning approach enables the model to learn nonlinear correlations between visual frames and word representations and to transfer its knowledge to words unseen during training. The experiments showed that the proposed method is capable of attaining very promising results on the most challenging publicly available dataset (LRS2), outperforming the two baselines by a large margin. Finally, we demonstrated its capacity in localizing the keyword in the frame sequence, even though we do not use any information about the location of the keyword during training.

\section{Acknowledgements}
This project has received funding from the European Union's Horizon 2020 research and innovation programme under the Marie Sklodowska-Curie grant agreement No  706668 (Talking Heads).
We are grateful to Dr. Stavros Petridis and Mr. Pingchuan Ma (i-bug, Imperial College London) for their contribution to the ASR-based experiments.
\newpage
\bibliographystyle{splncs}
\bibliography{AV}
\end{document}